\documentclass[10pt,journal,compsoc]{IEEEtran}
\ifCLASSOPTIONcompsoc
\usepackage[nocompress]{cite}
\else
\usepackage{cite}
\fi
\usepackage{amsmath}
\usepackage{graphicx}
\usepackage{subfigure}
\usepackage{subfig}
\usepackage[ruled,vlined,linesnumbered]{algorithm2e}
\usepackage{algorithmic}
\usepackage{url}
\usepackage{siunitx}
\usepackage{ragged2e}
\usepackage{array}
\usepackage{multirow}
\usepackage{multicol}
\usepackage{hhline}
\usepackage{amssymb}
\usepackage{ifpdf}
\usepackage{cite}
\usepackage{makecell}
\usepackage{placeins}

\ifCLASSINFOpdf
\else
\fi
\begin{document}
	%
	\title{Facial Landmark Point Localization using Coarse-to-Fine Deep Recurrent Neural Network}
	
	\author{Shahar~Mahpod,
		Rig~Das,
		Emanuele~Maiorana,
		Yosi~Keller,
		and~Patrizio~Campisi,
		\IEEEcompsocitemizethanks{\IEEEcompsocthanksitem S. Mahpod \& Y. Keller are with the Faculty of Engineering, Bar-Ilan University,
			Ramat Gan 52900, Israel (e-mail: \{mahpod.shahar, yosi.keller\}@gmail.com)
			\IEEEcompsocthanksitem R. Das, E. Maiorana \& P. Campisi are with the Section of Applied Electronics, Department of Engineering, Roma Tre University, Via Vito Volterra 62,
			00146, Rome, Italy (e-mail: \{rig.das, emanuele.maiorana, patrizio.campisi\}@uniroma3.it),
			Phone: +39.06.57337064, Fax: +39.06.5733.7026.\protect\\
		}
	}

	\markboth{Mahpod \MakeLowercase{\textit{et al.}}: CCNN}%
	{Shell \MakeLowercase{\textit{et al.}}: Bare Demo of IEEEtran.cls for Computer Society Journals}
	\IEEEtitleabstractindextext{%
		\justify
\begin{abstract}
	The accurate localization of facial landmarks is at the core of face
	analysis tasks, such as face recognition and facial expression analysis, to
	name a few. In this work we propose a novel localization approach based on a
	Deep Learning architecture that utilizes dual cascaded CNN subnetworks of
	the same length, where each subnetwork in a cascade refines the accuracy of
	its predecessor. The first set of cascaded subnetworks estimates heatmaps
	that encode the landmarks' locations, while the second set of cascaded
	subnetworks refines the heatmaps-based localization using regression, and
	also receives as input the output of the corresponding heatmap estimation
	subnetwork. The proposed scheme is experimentally shown to compare favorably
	with contemporary state-of-the-art schemes.
\end{abstract}

		\begin{IEEEkeywords}
	Face Alignment, Facial Landmark Localization, Convolutional Neural Networks, Deep Cascaded Neural Networks.
	\end{IEEEkeywords}}

	\maketitle

	\IEEEdisplaynontitleabstractindextext

	%
	\IEEEpeerreviewmaketitle

	\IEEEraisesectionheading{\section{Introduction}\label{Introduction}}

\IEEEpeerreviewmaketitle

The localization of facial landmark points, such as eyebrows, eyes, nose,
mouth and jawline, is one of the core computational components in visual
face analysis, and is applied in a gamut of applications: face recognition 
\cite{Huang13}, face verification \cite{Lu2015}, and facial attribute
inference \cite{NKumar08}, to name a few. Robust and accurate localization
entails difficulties due to varying face poses, illumination, resolution
variations, and partial occlusions, as depicted in Fig. \ref%
{fig:FacialLandmark}. 
\begin{figure}[tbh]
	\subfigure[]{\includegraphics[width=0.475\linewidth]{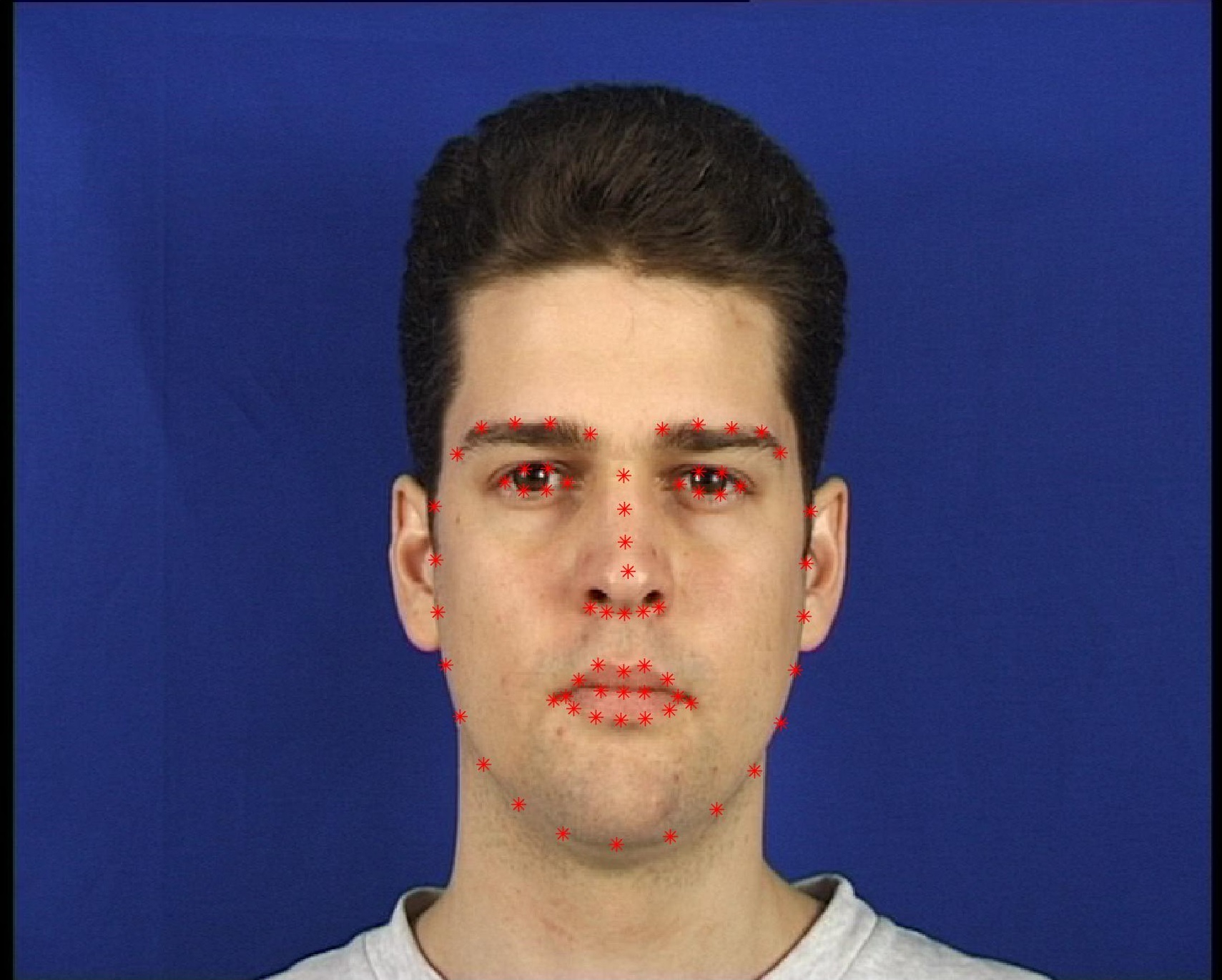}} %
	\subfigure[]{\includegraphics[width=0.475\linewidth]{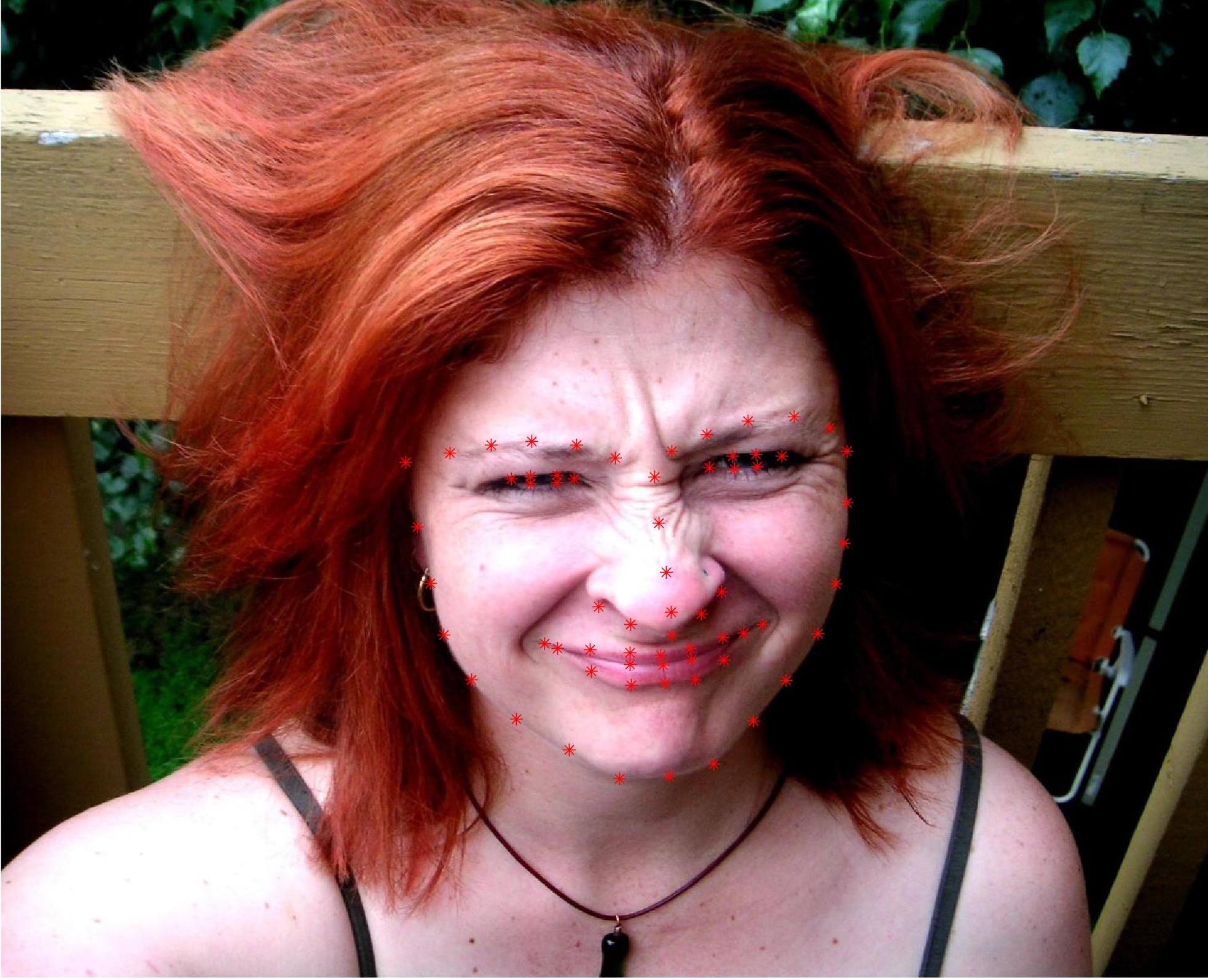}}
	\caption{Facial landmark localization. Each image feature, marked by a point
		is considered a particular landmark and is localized individually. (a) A
		frontal face image from the XM2VTS datasets \protect\cite{Messer03}. (b) An
		image from the Helen dataset \protect\cite{Le2012} characterized by a
		non-frontal pose and expression variation, making the localization
		challenging.}
	\label{fig:FacialLandmark}
\end{figure}

Classical face localization schemes such as Active Appearance Models (AAM) 
\cite{Cootes01} and Active shape models (ASM) \cite{Cootes92} apply
generative models aiming to learn a parametric statistical model of the face
shape and its gray-level appearance in a training phase. The model is
applied in test time to minimize the residual between the training image and
the synthesized model. Parametric shape models such as Constrained Local
Models (CLM) \cite{Cristinacce06}, utilize Bayesian formulations for shape
constrained searches, to estimate the landmarks iteratively. Nonparametric
global shape models \cite{Belhumeur13} apply SVM classification to estimate
the landmarks under significant appearance changes and large pose variations.

Regression based approaches \cite{Cao12,Xiong13} learn high dimensional
regression models that iteratively estimate landmarks positions using local
image features, and showed improved accuracy when applied to in-the-wild
face images. Such schemes are initiated using an initial estimate and are in
general limited to yaw, pitch and head roll angles of less than $30%
{{}^\circ}%
$.

Following advances in object detection, parts-based models were applied to
face localization \cite{Zhu12,Zhu15} where the facial landmarks and their
geometric relationships are encoded by graphs. Computer vision was
revolutionized by Deep Learning-based approaches that were also applied to
face localization \cite{RRanjan17,Zhang16,Kowalski17}, yielding robust and
accurate estimates. Convolutional neural networks (CNNs) extract high level
features over the whole face region and are trained to predict all of the
keypoints simultaneously, while avoiding local minima. In particular,
heatmaps were used in CNN-based landmark localization schemes following the
seminal work of Pfister et al. \cite{pfister2015flowing}, extended by the
iterative formulation by Belagiannis and Zisserman \cite{7961778}.

\begin{figure}[tbh]
	\centering\includegraphics[width=0.5\textwidth]{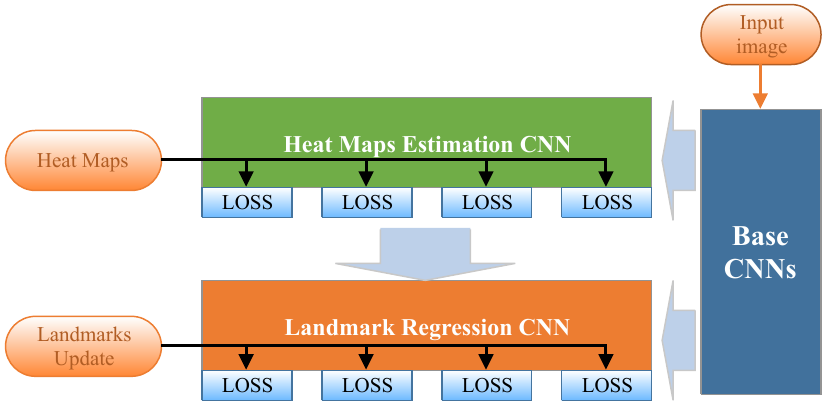}
	\caption{The outline of the proposed CCNN framework. The CCNN consists of
		Base CNNs that are proceeded by the Cascaded Heatmap subnetwork (CHCNN) that
		estimates the heatmaps and the Cascaded Regression CNN (CRCNN) that refines
		the heatmaps localization via pointwise regression.}
	\label{fig:general CCNN}
\end{figure}

In this work we propose a novel Deep Learning-based framework for facial
landmark localization that is formulated as a Cascaded CNN (CCNN) consisting
of dual cascaded heatmaps and regression subnetworks. A outline of the
architecture of the proposed CNN is depicted in Fig. \ref{fig:general CCNN},
where after computing the feature maps of the entire image, each facial
landmark is coarsely localized by a particular heatmap, and all
localizations are refined by regression subnetworks. In that we extend prior
works \cite{pfister2015flowing,7961778} where iterative heatmaps
computations were used, without the additional refinement subnetwork
proposed in our work. The heatmaps are estimated using a Cascaded Heatmap
subnetwork (CHCNN) consisting of multiple successive heatmap-based
localization subnetworks, that compute a coarse-to-fine estimate of the
landmark localization. This localization estimate is refined by applying the
Cascaded Regression CNN (CRCNN) subnetwork. The cascaded layers in both the
CHCNN and CRCNN are non-weight-sharing, allowing to separately learn a
particular range of localizations. The CCNN is experimentally shown to
compare favourably with contemporary state-of-the-art face localization
schemes. Although this work exemplifies the use of the proposed approach in
the localization of facial landmarks, it is of general applicability and can
be used for any class of objects, given an appropriate annotated training
set. 

Thus, we propose the following contributions:

\textbf{First}, we derive a face localizations scheme based on CNN-based
heatmaps estimation and refinement by a corresponding regression CNN.

\textbf{Second}, both heatmap estimation and regression are formulated as
cascaded subnetworks that allow iterative refinement of the localization
accuracy. To the best of our knowledge, this is the first such formulation
for the face localization problem.

\textbf{Last}, the proposed CCNN framework is experimentally shown to
outperform contemporary state-of-the-art approaches.

This paper is organized as follows: Section \ref{SOA_Face_Landmark} provides
an overview of the state-of-the-art techniques for facial landmark
localization, while Section \ref{Model} introduces the proposed CCNN and its
CNN architecture. The experimental validation and comparison to
state-of-the-art methods is detailed in Section \ref{results}. Conclusions
are drawn in Section \ref{conclusion}

\section{Related work}

\label{SOA_Face_Landmark}

The localization of facial landmarks, being a fundamental computer vision
task, was studied in a multitude of works, dating back to the seminal
results in Active Appearance Models (AAM) \cite{Cootes01} and Constrained
Local Models (CLM) \cite{Cristinacce06} that paved the way for recent
localization schemes. In particular, the proposed scheme relates to the
Cascaded Shape Regression (CSR), \cite{Trigeorgis16} and Deep Learning-based 
\cite{Zhang16,RRanjan17,Xiao16,Lai17,Shao17} models.

CSR schemes localize the landmark points explicitly by iterative regression,
where the regression estimates the localization refinement offset using the
local image features computed at the estimated landmarks locations. Such
schemes are commonly initiated by an initial estimate of the landmarks based
on an average face template, and a bounding box of the face detected by a
face detector, such as Viola-Jones \cite{Viola01}. Thus, the Supervised
Descent Method by Xiong and De-la-Torre \cite{Xiong13} learned a cascaded
linear regression using SIFT features \cite{Lowe04} computed at the
estimated landmark locations. Other schemes strived for computational
efficiency by utilizing Local Binary Features (LBF) that are learnt by
binary trees in a training phase. Thus, Ren et al. in \cite{Ren14} proposed
a face alignment technique achieving 3000 fps by learning highly
discriminative LBFs for each facial landmark independently, and the learned
LBFs\ are used to jointly learn a linear regression to estimate the facial
landmarks' locations.

Chen et al. \cite{Chen14} applied random regression forests to landmark
localization using Haar-like local image features to achieve computational
efficiency. Similarly, a discriminative regression approach was proposed by\
Asthana et al. \cite{Asthana13} to learn regression functions from the image
encodings to the space of shape parameters. A cascade of a mixture of
regressors was suggested by Tuzel et al. \cite{Tuzel16}, where each
regressor learns a regression model adapted to a particular subspace of pose
and expressions, such as a smiling face turned to the left. Affine
invariance was achieved by aligning each face to a canonical shape before
applying the regression.

A parts-based approach for a unified approach to face detection, pose
estimation, and landmark localization was suggested by Zhu and Ramanan \cite%
{Zhu12}, where the facial features and their geometrical relations are
encoded by the vertices of a corresponding graph. The inference is given by
a mixture of trees trained using a training set. An iterative coarse-to-fine
refinement implemented in space-shape was introduced by Zhu et al. \cite%
{Zhu15}, where the initial coarse solution allows to constrain the search
space of the finer shapes. This allows to avoid suboptimal local minima and
improves the estimation of large pose variations.

Deep Learning was also applied to face alignment by extending
regression-based schemes for face alignment. The Mnemonic Descent Method by
Trigeorgis et al. \cite{Trigeorgis16} combines regression as in CSR schemes,
with feature learning using Convolutional Neural Networks (CNNs). The image
features are learnt by the convolution layers, followed by a cascaded neural
network that is jointly trained, yielding an end-to-end trainable scheme.

Autoencoders were applied by Zhang et al. \cite{Zhang14} in a coarse-to-fine
scheme, using successive stacked autoencoders. The first subnetwork predicts
an initial estimate of the landmarks utilizing a low-resolution input image.
The following subnetworks progressively refine the landmarks' localization
using the local features extracted around the current landmarks. A similar
CNN-based approach was proposed by Shi et al. \cite{Shi18}, where the
subnetworks were based on CNNs, and a coarse face shape is initially
estimated, while the following layers iteratively refine the face landmarks.
CNNs were applied by Zhou et al. \cite{Zhou13} to iteratively refine a
subset of facial landmarks estimated by preceding network layers, where each
layer predicts the position and rotation angles of each facial feature. Xiao
et al. in \cite{Xiao16} introduced a cascaded localization CNN using
cascaded regressions that refine the localization progressively. The
landmark locations are refined sequentially at each stage, allowing the more
reliable landmark points to be refined earlier, where LSTMs are used to
identify the reliable landmarks and refine their localization.

A conditional Generative Adversarial Network (GAN) was applied by Chen et
al. \cite{Chen_arxiv17} to face localization to induce geometric priors on
the face landmarks, by introducing a discriminator that classifies real vs.
erroneous (\textquotedblleft fake\textquotedblright ) localizations. A CNN
with multiple losses was derived by Ranjan et al. in \cite{RRanjan17} for
simultaneous face detection, landmarks localization, pose estimation and
gender recognition. The proposed method utilizes the lower and intermediate
layers of the CNN followed by multiple subnetworks, each with a different
loss, corresponding to a particular tasks, such as face detection etc.
Multi-task estimation of multiple facial attributes, such as gender,
expression, and appearance attributes was also proposed by Zhang et al. \cite%
{Zhang16}, and was shown to improve the estimation robustness and accuracy.

Multi-task CNNs with auxiliary losses were applied by Sina et al. \cite%
{Sina17} for training a localization scheme using partially annotated
datasets where accurate landmark locations are only provided for a small
data subset, but where class labels for additional related cues are
available. They propose a sequential multitasking scheme where the class
labels are used via auxiliary losses. An unsupervised landmark localization
scheme is also proposed, where the model is trained to produce equivalent
landmark locations with respect to a set of transformations that are applied
to the image.

Pfister et al. \cite{pfister2015flowing} introduced the use of heatmaps for
landmark localization by CNN-based formulation. It was applied to human pose
estimation in videos where the landmark points marks body parts, and optical
flow was used to fuse heatmap predictions from neighboring frames. This
approach was extended by Belagiannis and Zisserman by deriving a cascaded
heatmap estimation subnetwork, consisting of multiple heatmap regression
units, where the heatmap is estimated progressively such that each heatmap
regression units received as input its predecessor's output. This school of
thought is of particular interest to our work that is also heatmaps-based,
but also applies a cascaded regression subnetwork that refines the heatmap
estimate.

Bulat and Tzimiropoulos \cite{Bulat16} applied convolutional heatmap
regression to 3D face alignment, by estimating the 2D coordinates of the
facial landmarks using a set of 2D heatmaps, one per landmark, estimated
using a CNN\ with an $L_{2}$ regression loss. Another CNN is applied to the
estimated heatmaps and the input RGB image to estimate the $Z$ coordinate. A
scheme consisting of two phases was proposed by Shao et al. \cite{Shao17}
where the image features are first estimated by a heatmap, and then refined
by a set of shape regression subnetworks each adapted and trained for a
particular pose.

Kowalski et al. \cite{Kowalski17} proposed a multistage scheme for face
alignment. It is based on a cascaded CNN where each stage refines the
landmark positions estimated at the previous one. The inputs to each stage
are a face image normalized to a canonical pose, the features computed by
the previous stage, and a heatmap computed using the results of the previous
phase. The heatmap is not estimated by the CNN, and in that, this scheme
differs significantly from the proposed scheme, and other schemes that
directly estimate the heatmaps as part of the CNN \cite%
{pfister2015flowing,Shao17,Bulat16}. 
\begin{table}[tbh]
	\caption{Overview of contemporary state-of-the-art facial landmark
		localization schemes.}
	\label{tab:overview}{\tiny \ \centering  \renewcommand{\arraystretch}{2.0} 
		\resizebox{9cm}{!}{
			\begin{tabular}{|c|c|c|c|c|}
				\hline             \fontsize{10pt}{20pt}\selectfont\textbf{Paper}&\fontsize{10pt}{20pt}\selectfont\textbf{Training Sets}
				&\fontsize{10pt}{20pt}\selectfont\textbf{Test Sets}
				&\fontsize{10pt}{20pt}\selectfont\textbf{Pts\#}
				\\
				\hline
				\multirow{3}{*}{\fontsize{10pt}{20pt}\selectfont Ren \cite{Ren14}}
				&\fontsize{10pt}{20pt}\selectfont LFPW\cite{Belhumeur13},
				&\fontsize{10pt}{20pt}\selectfont i-bug\cite{Sagonas16}    &\multirow{3}{*}{\fontsize{10pt}{20pt}\selectfont  68}
				\\
				\hhline{~~-~}
				&\fontsize{10pt}{20pt}\selectfont Helen\cite{Le2012},  &\fontsize{10pt}{20pt}\selectfont Helen+LFPW\cite{Le2012,Belhumeur13}
				&
				\\
				\hhline{~~-~}
				&\fontsize{10pt}{20pt}\selectfont AFW,300-W
				&\fontsize{10pt}{20pt}\selectfont i-bug+Helen+LFPW
				&
				\\
				\hline			
				\multirow{4}{*}{\fontsize{10pt}{20pt}\selectfont Ranjan \cite{RRanjan17}}
				&\multirow{4}{*}{\fontsize{10pt}{20pt}\selectfont AFLW\cite{koestinger11}}     &\multirow{2}{*}{\fontsize{10pt}{20pt}\selectfont i-bug}    &\multirow{2}{*}{\fontsize{10pt}{20pt}\selectfont  68}
				\\
				\hhline{~~~~}
				&&& \\
				\hhline{~~--}
				&&\multirow{2}{*}{\fontsize{10pt}{20pt}\selectfont AFLW\cite{koestinger11}}
				&\multirow{2}{*}{\fontsize{10pt}{20pt}\selectfont  21}
				
				\\
				\hhline{~~~~}
				&&&			
				\\
				\hline
				\multirow{5}{*}{\fontsize{10pt}{20pt}\selectfont Zhu \cite{Zhu15}}
				&\fontsize{10pt}{20pt}\selectfont LFPW, &\fontsize{10pt}{20pt}\selectfont i-bug
				&\multirow{5}{*}{\fontsize{10pt}{20pt}\selectfont  68}
				\\
				\hhline{~~-~}
				&\fontsize{10pt}{20pt}\selectfont Helen,  &\fontsize{10pt}{20pt}\selectfont Helen+LFPW
				& \\
				\hhline{~~-~}
				&\fontsize{10pt}{20pt}\selectfont AFW,
				&\fontsize{10pt}{20pt}\selectfont i-bug+Helen+LFPW
				& \\
				\hhline{~~-~}
				&\fontsize{10pt}{20pt}\selectfont 300-W
				&\fontsize{10pt}{20pt}\selectfont LFPW\cite{Belhumeur13}
				&  \\
				\hhline{~~-~}
				&\fontsize{10pt}{20pt}\selectfont
				&\fontsize{10pt}{20pt}\selectfont Helen\cite{Le2012}
				&  \\
				\hline			
				\multirow{4}{*}{\fontsize{10pt}{20pt}\selectfont Zhang \cite{Zhang16}}
				&\fontsize{10pt}{20pt}\selectfont MAFL\cite{Sun14}, &\fontsize{10pt}{20pt}\selectfont i-bug &\multirow{4}{*}{\fontsize{10pt}{20pt}\selectfont  68}
				\\
				\hhline{~~-~}
				&\fontsize{10pt}{20pt}\selectfont AFLW,  &\fontsize{10pt}{20pt}\selectfont Helen+LFPW
				&  \\
				\hhline{~~-~}
				&\fontsize{10pt}{20pt}\selectfont COFW\cite{Burgos13},
				& \fontsize{10pt}{20pt}\selectfont i-bug+Helen+LFPW
				&
				\\
				\hhline{~~-~}
				&\fontsize{10pt}{20pt}\selectfont Helen,300-W
				&\fontsize{10pt}{20pt}\selectfont Helen
				&
				\\
				\hline
				\multirow{5}{*}{\fontsize{10pt}{20pt}\selectfont Xiao \cite{Xiao16}}
				&\fontsize{10pt}{20pt}\selectfont LFPW, &\fontsize{10pt}{20pt}\selectfont i-bug    &\multirow{5}{*}{\fontsize{10pt}{20pt}\selectfont  68}
				\\
				\hhline{~~-~}
				&\fontsize{10pt}{20pt}\selectfont Helen,  &\fontsize{10pt}{20pt}\selectfont Helen+LFPW
				& \\
				\hhline{~~-~}
				&\fontsize{10pt}{20pt}\selectfont AFW,
				&\fontsize{10pt}{20pt}\selectfont i-bug+Helen+LFPW
				& \\
				\hhline{~~-~}
				&\fontsize{10pt}{20pt}\selectfont 300-W
				&\fontsize{10pt}{20pt}\selectfont LFPW
				&  \\
				\hhline{~~-~}
				&\fontsize{10pt}{20pt}\selectfont
				&\fontsize{10pt}{20pt}\selectfont Helen
				&  \\
				\hline
				\multirow{5}{*}{\fontsize{10pt}{20pt}\selectfont Lai \cite{Lai17}}
				&\fontsize{10pt}{20pt}\selectfont LFPW, &\fontsize{10pt}{20pt}\selectfont i-bug    &\multirow{5}{*}{\fontsize{10pt}{20pt}\selectfont  68}
				\\
				\hhline{~~-~}
				&\fontsize{10pt}{20pt}\selectfont Helen,  &\fontsize{10pt}{20pt}\selectfont Helen+LFPW
				& \\
				\hhline{~~-~}
				&\fontsize{10pt}{20pt}\selectfont AFW,
				&\fontsize{10pt}{20pt}\selectfont i-bug+Helen+LFPW
				&  \\
				\hhline{~~-~}
				&\fontsize{10pt}{20pt}\selectfont 300-W
				&\fontsize{10pt}{20pt}\selectfont LFPW
				&   \\
				\hhline{~~-~}
				&\fontsize{10pt}{20pt}\selectfont
				&\fontsize{10pt}{20pt}\selectfont Helen
				&   \\
				\hline		
				\multirow{3}{*}{\fontsize{10pt}{20pt}\selectfont Shao \cite{Shao17}}
				&\fontsize{10pt}{20pt}\selectfont CelebA\cite{Liu15}, &\fontsize{10pt}{20pt}\selectfont i-bug     &\multirow{3}{*}{\fontsize{10pt}{20pt}\selectfont  68}
				\\
				\hhline{~~-~}
				&\fontsize{10pt}{20pt}\selectfont 300-W,  &\fontsize{10pt}{20pt}\selectfont Helen+LFPW
				&
				\\
				\hhline{~~-~}
				&\fontsize{10pt}{20pt}\selectfont MENPO\cite{Zafeiriou17}
				&\fontsize{10pt}{20pt}\selectfont i-bug+Helen+LFPW
				&
				\\
				\hline
				\multirow{3}{*}{\fontsize{10pt}{20pt}\selectfont Sina \cite{Sina17}}
				&\fontsize{10pt}{20pt}\selectfont Helen, &\fontsize{10pt}{20pt}\selectfont i-bug    &\multirow{3}{*}{\fontsize{10pt}{20pt}\selectfont  68}
				\\
				\hhline{~~-~}
				&\fontsize{10pt}{20pt}\selectfont AFW,  &\fontsize{10pt}{20pt}\selectfont Helen+LFPW
				&
				\\
				\hhline{~~-~}
				&\fontsize{10pt}{20pt}\selectfont LFPW
				&\fontsize{10pt}{20pt}\selectfont i-bug+Helen+LFPW
				&
				\\
				\hline
				\multirow{3}{*}{\fontsize{10pt}{20pt}\selectfont Chen \cite{Chen17}}
				&\fontsize{10pt}{20pt}\selectfont Helen, &\fontsize{10pt}{20pt}\selectfont i-bug    &\multirow{3}{*}{\fontsize{10pt}{20pt}\selectfont  68}
				\\
				\hhline{~~-~}
				&\fontsize{10pt}{20pt}\selectfont 300-W,  &\fontsize{10pt}{20pt}\selectfont Helen+LFPW
				&
				\\
				\hhline{~~-~}
				&\fontsize{10pt}{20pt}\selectfont MENPO
				&\fontsize{10pt}{20pt}\selectfont i-bug+Helen+LFPW
				&
				\\
				\hline
				
				\multirow{8}{*}{\fontsize{10pt}{20pt}\selectfont Kowalski \cite{Kowalski17}}
				&\fontsize{10pt}{20pt}\selectfont LFPW,
				&\fontsize{10pt}{20pt}\selectfont i-bug   &\multirow{8}{*}{\fontsize{10pt}{20pt}\selectfont  68}
				\\
				\hhline{~~-~}
				&\fontsize{10pt}{20pt}\selectfont Helen, &\fontsize{10pt}{20pt}\selectfont Helen+LFPW &  \\
				\hhline{~~-~}
				&\fontsize{10pt}{20pt}\selectfont AFW, &\fontsize{10pt}{20pt}\selectfont i-bug+Helen+LFPW &
				\\
				\hhline{~~-~}
				&\fontsize{10pt}{20pt}\selectfont 300-W
				&\fontsize{10pt}{20pt}\selectfont 300-W private test set &
				\\
				\hhline{~--~}
				&\fontsize{10pt}{20pt}\selectfont LFPW,
				&\fontsize{10pt}{20pt}\selectfont i-bug   &
				\\
				\hhline{~~-~}
				&\fontsize{10pt}{20pt}\selectfont Helen,AFW, &\fontsize{10pt}{20pt}\selectfont Helen+LFPW & \\
				\hhline{~~-~}
				&\fontsize{10pt}{20pt}\selectfont 300-W, &\fontsize{10pt}{20pt}\selectfont i-bug+Helen+LFPW &
				\\
				\hhline{~~-~}
				&\fontsize{10pt}{20pt}\selectfont  MENPO &\fontsize{10pt}{20pt}\selectfont 300-W private test set &
				\\
				\hline

\multirow{4}{*}{\fontsize{10pt}{20pt}\selectfont He \cite{He17}}
&\fontsize{10pt}{20pt}\selectfont LFPW,

&\multirow{4}{*}{\fontsize{10pt}{20pt}\selectfont i-bug}
  &\multirow{4}{*}{\fontsize{10pt}{20pt}\selectfont  68}
\\
&\fontsize{10pt}{20pt}\selectfont Helen,AFW &&  \\
&\fontsize{10pt}{20pt}\selectfont 300-W, & &
\\
&\fontsize{10pt}{20pt}\selectfont MENPO
&&
\\
\hline

	\multirow{3}{*}{\fontsize{10pt}{20pt}\selectfont Chen \cite{Chen_arxiv17}}
	&\fontsize{10pt}{20pt}\selectfont LFPW,
	
	&\multirow{3}{*}{\fontsize{10pt}{20pt}\selectfont 300-W private test set}
	&\multirow{3}{*}{\fontsize{10pt}{20pt}\selectfont  68}
	\\
	&\fontsize{10pt}{20pt}\selectfont Helen,AFW &&  \\
	&\fontsize{10pt}{20pt}\selectfont i-bug, & &
	\\
	\hline

			\end{tabular}
	} }
\end{table}
Last, we summarize the different face localization approaches in Table \ref%
{tab:overview}, where we detail the training and test datasets, as these are
the basis for forming the experimental validation in Section \ref{results}.

\section{Face Localization using cascaded CNNs}

\label{Model}

The face localization problem is the localization of a set of landmarks
points ${P}=\left\{ \mathbf{p}_{i}\right\} _{1}^{N}$, such that $\mathbf{p}%
_{i}$=$[x_{i},y_{i}]^{T},$ in a face image $I\in \mathbb{R}^{w\times h\times
	3}$. The number of estimated points $N$ relates to the annotation convention
used, and in this work we used $N=68$ landmark points following most
contemporary works. The general and detailed outlines of the proposed CCNN's
architecture are depicted in Figs. \ref{fig:general CCNN} and \ref%
{fig:details CCNN}, respectively. It comprises of three subnetworks, where
the first is a pseudo-siamese (non-weight-sharing) subnetwork consisting of
two subnetworks $\left\{ BaseCNN_{1},BaseCNN_{2}\right\} $ that compute the
corresponding feature maps $\left\{ \mathbf{F}_{1},\mathbf{F}_{2}\right\} $
of the input image and an initial estimate of the heatmaps. 
\begin{figure}[tbh]
	\centering
	\includegraphics[width=0.5\textwidth]{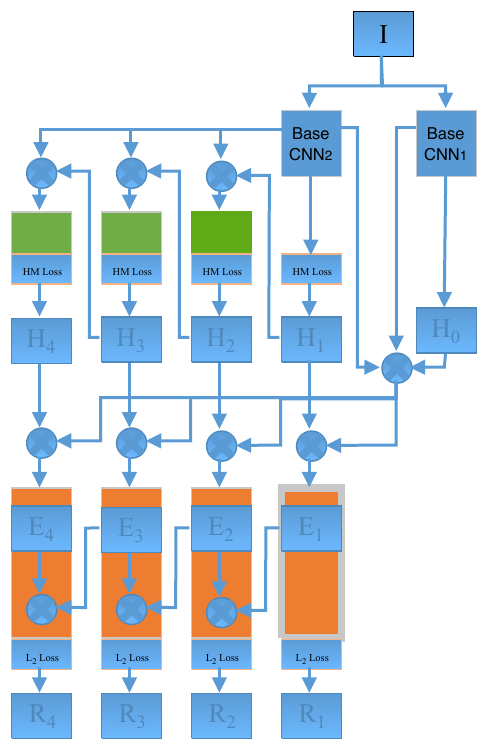}
	\caption{A schematic visualization of the proposed CCNN localization
		network. The input image is analyzed by the two Base subnetworks $\left\{
		BaseCNN_{1},BaseCNN_{2}\right\} $, and the Cascaded Heatmap CNN\ (CHCNN)
		consisting of four heatmap (HM) estimation units $H_{1}-H_{4}$. Their
		results are refined by the Cascaded Regression CNN\ (CHCNN) consisting of
		four regression units $R_{1}-R_{4}$. The symbol $\oplus $ relates to the
		concatenation of variables along their third dimension.}
	\label{fig:details CCNN}
\end{figure}

The second subnetwork is the cascaded heatmap subnetwork (CHCNN) that
robustly estimates the heatmaps, that encode the landmarks, a single 2D\
heatmap per facial feature location. The heatmaps are depicted in Fig. \ref%
{fig:heatmaps}. The CHCNN consists of $K=4$ cascaded 3D heatmaps estimation
units, detailed in Section \ref{Heat-map_cascaded_level}, that estimate $K$
3D heatmaps $\{\mathbf{H}_{k}\}_{1}^{K}$ such that $\mathbf{H}_{k}\in 
\mathbb{R}^{64\times 64\times N}$. The cascaded formulation implies that
each CHCNN subunit $\mathbf{H}_{k}$ is given as input the heatmap estimated
by its preceding subunit $\mathbf{H}_{k-1}$, alongside the feature map $%
\mathbf{F}_{1}$. The heatmap subunits are non-weight-sharing, as each
subunit refines a different estimate of the heatmaps. In that, the proposed
schemes differs from the heatmaps-based pose estimation of Belagiannis and
Zisserman \cite{7961778} that applies weight-sharing cascaded units. The
output of the CHCNN are the locations of the maxima of $\{\mathbf{H}%
_{k}\}_{1}^{K}$ denoted $\{\widehat{\mathbf{P}}_{k}\}_{1}^{K}$, such that $%
\widehat{\mathbf{P}}_{k}=\left\{ \widehat{\mathbf{p}}_{k}^{i}\right\} $.

As the heatmap-based estimates $\{\widehat{\mathbf{P}}_{k}\}_{1}^{K}$ are
given on a coarse grid, their locations are refined by applying the Cascaded
Regression CNN (CRCNN) detailed in Section \ref{Heat-map_cascaded_level}.
The CRCNN consists of $K$ cascaded regression subunits $\{\mathbf{E}%
_{k}\}_{1}^{K}$, where each regression subunit $\mathbf{E}_{k}$ applies a
regression loss to refine the corresponding heatmaps-based landmark estimate 
$\widehat{\mathbf{P}}_{k}$, and estimate the refinement $\Delta \widehat{%
	\mathbf{P}}_{k}$ 
\begin{equation}
\Delta \widehat{\mathbf{P}}_{k}=vec\left( P_{k}\right) -vec\left( \widehat{%
	\mathbf{P}}_{k}\right) ,  \label{equ:regression}
\end{equation}%
where $vec\left( \cdot \right) $ is a vectorized replica of the $N$ points
in a set, and Eq. \ref{equ:regression} is optimized using an $L_{2}$ loss. 
\begin{figure*}[tbh]
	\includegraphics[width=\textwidth]{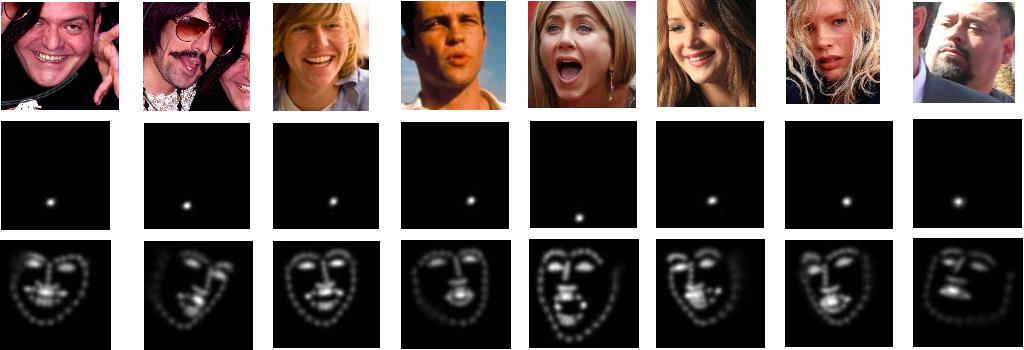}
	\caption{Visualizations of facial landmarks localization heatmaps. The first
		row shows the face images, while the second row depicts a corresponding
		single heatmap of a particular facial feature. The third row shows the
		corresponding $N=68$ points of all heatmap.}
	\label{fig:heatmaps}
\end{figure*}

\subsection{Base subnetwork}

\label{Base_Blocks}

The Base subnetwork consists of two pseudo-siamese (non-weight-sharing)
subnetworks detailed in Table \ref{tab:Base_block}. The first part of the
subnetwork, layers A1-A7 in Table \ref{tab:Base_block}, computes the feature
maps of the input image. The succeeding layers A8-A12 compute an estimate of
the $N=68$ heatmaps, one per facial feature. These layers apply filters with
wide $9\times 9$ support to encode the relations between neighboring facial
features.

The base CNNs $\left\{ BaseCNN_{1},BaseCNN_{2}\right\} $ and corresponding
feature maps $\left\{ \mathbf{F}_{1},\mathbf{F}_{2}\right\} $ are trained
using different losses and backpropagation paths as depicted in Fig. \ref%
{fig:details CCNN}. $BaseCNN_{1}$ is connected to the CRCNN and is thus
adapted to the regression task, while $BaseCNN_{2}$ is connected to the
CHCNN and its feature map $\mathbf{F}_{2}$ is adapted to the heatmaps
estimation task. $BaseCNN_{2}$ computes the initial estimate of the heatmaps
and is trained using a $L_{2}$ loss, while $BaseCNN_{1}$ is trained using
the backpropagation of the CRCNN subnetwork. 
\begin{table}[tbh]
	\caption{Base subnetwork architecture. Given the input image, the base
		subnetwork estimates the feature map $F$ and the heatmap $H$. Two such
		non-weight-sharing subnetworks are used as depicted in Fig. \protect\ref%
		{fig:details CCNN}.}
	\label{tab:Base_block}{\tiny \ \centering
	}
	\par
	\begin{tabular}{|l|c|c|l|c|}
		\hline
		$\mathbf{Layer}$ & \textbf{Feature Map} & $\mathbf{F_{Size}}$ & \textbf{%
			Stride} & \textbf{Pad} \\ \hline\hline
		\textbf{Input : }$I$ & 256 x 256 x 3 & - & - & - \\ \hline
		A1-Conv & 256 x 256 x 3 & 3 x 3 & 1 x 1 & 2 x 2 \\ \hline
		A1-ReLu & 256 x 256 x 64 & - & - & - \\ \hline
		A2-Conv & 256 x 256 x 64 & 3 x 3 & 1 x 1 & 2 x 2 \\ \hline
		A2-ReLu & 256 x 256 x 64 & - & - & - \\ \hline
		A2-Pool & 256 x 256 x 64 & 2 x 2 & 2 x 2 & 0 x 0 \\ \hline
		A3-Conv & 128 x 128 x 64 & 3 x 3 & 1 x 1 & 2 x 2 \\ \hline
		A3-ReLu & 128 x 128 x 64 & - & - & - \\ \hline
		A4-Conv & 128 x 128 x 64 & 3 x 3 & 1 x 1 & 2 x 2 \\ \hline
		A4-ReLu & 128 x 128 x 128 & - & - & - \\ \hline
		A4-Pool & 128 x 128 x 128 & 2 x 2 & 2 x 2 & 0 x 0 \\ \hline
		A5-Conv & 64 x 64 x 128 & 3 x 3 & 1 x 1 & 2 x 2 \\ \hline
		A5-ReLu & 64 x 64 x 128 & - & - & - \\ \hline
		A6-Conv & 64 x 64 x 128 & 3 x 3 & 1 x 1 & 2 x 2 \\ \hline
		A6-ReLu & 64 x 64 x 128 & - & - & - \\ \hline
		A7-Conv & 64 x 64 x 128 & 1 x 1 & 1 x 1 & - \\ \hline
		\textbf{Output : }$F$ & 64 x 64 x 128 & - & - & - \\ \hline
		A8-Conv & 64 x 64 x 128 & 9 x 9 & 1 x 1 & 8 x 8 \\ \hline
		A8-ReLu & 64 x 64 x 128 & - & - & - \\ \hline
		A9-Conv & 64 x 64 x 128 & 9 x 9 & 1 x 1 & 8 x 8 \\ \hline
		A9-ReLu & 64 x 64 x 128 & - & - & - \\ \hline
		A10-Conv & 64 x 64 x 128 & 1 x 1 & 1 x 1 & 0 x 0 \\ \hline
		A10-ReLu & 64 x 64 x 256 & - & - & - \\ \hline
		A11-Conv & 64 x 64 x 256 & 1 x 1 & 1 x 1 & 0 x 0 \\ \hline
		A11-ReLu & 64 x 64 x 256 & - & - & - \\ \hline
		A11-Dropout0.5 & 64 x 64 x 256 & - & - & - \\ \hline
		A12-Conv & 64 x 64 x 256 & 1 x 1 & 1 x 1 & 0 x 0 \\ \hline
		A12-ReLu & 64 x 64 x 68 & - & - & - \\ \hline
		\textbf{Output : }$H$ & 64 x 64 x 68 & - & - & - \\ \hline
	\end{tabular}%
\end{table}

\subsection{Cascaded heatmap estimation CNN}

\label{Heat-map_cascaded_level}

The heatmap images encode the positions of the set of landmarks points ${P}%
=\left\{ \mathbf{p}_{i}\right\} _{1}^{N}$, by relating a single heatmap $%
\mathbf{H}^{i}$ per landmark $\mathbf{p}_{i}$ to the location of the maximum
of $\mathbf{H}^{i}$, where the heatmaps are compute in a coarse resolution
of $1/4$ of the input image resolution. The heatmaps are computed using the
CHCNN subnetwork consisting of $K=4$ Heatmap Estimation Subunits (HMSU)
detailed in Table \ref{tab:heatmap_cascaded_branch}.

The cascaded architecture of the CHCNN implies that each heatmap subunit
estimates a heatmap $\mathbf{H}_{k}$ and receives as input the heatmap $%
\mathbf{H}_{k-1}$ estimated by the previous subunit, and a feature map $%
\mathbf{F}_{2}$ estimated by the Base subnetwork $BaseCNN_{2}$. The
different inputs are concatenated as channels such that the input is given
by $\mathbf{F}_{2}\oplus \mathbf{H}_{k-1}$.

The HMSU architecture comprises of wide filters, $[7\times 7]$ and $%
[13\times 13]$ corresponding to layers B1 and B2, respectively, in Table \ref%
{tab:heatmap_cascaded_branch}. These layers encode the geometric
relationships between relatively distant landmarks. Each heatmap is trained
with respect to a $L_{2}$ loss, and in the training phase, the locations of
the facial landmarks are labeled by narrow Gaussians centered at the
landmark location, to improve the training convergence. 
\begin{table}[tbh]
	\caption{The heatmap estimation subunit. The heatmap CNN (CHCNN) is a
		cascaded CNN consisting of a series of $K=4$ subunits. The input to each
		subunit is the output of the previous subunit and the feature map $F_{2}$.}
	\label{tab:heatmap_cascaded_branch}\centering
	\par
	\begin{tabular}{|l|c|c|l|c|}
		\hline
		$\mathbf{L_{Type}}$ & \textbf{Feature Map} & $\mathbf{F_{Size}}$ & \textbf{%
			Stride} & \textbf{Pad} \\ \hline\hline
		\textbf{Input : } $F_{1}\oplus {H_{k-1}}$ & 64 x 64 x 136 & - & - & - \\ 
		\hline
		B1-Conv & 64 x 64 x 136 & 7 x 7 & 1 x 1 & 6 x 6 \\ \hline
		B1-ReLu & 64 x 64 x 64 & - & - & - \\ \hline
		B2-Conv & 64 x 64 x 64 & 13 x 13 & 1 x 1 & 12 x 12 \\ \hline
		B2-ReLu & 64 x 64 x 64 & - & - & - \\ \hline
		B3-Conv & 64 x 64 x 64 & 1 x 1 & 1 x 1 & 0 x 0 \\ \hline
		B3-ReLu & 64 x 64 x 128 & - & - & - \\ \hline
		B4-Conv & 64 x 64 x 128 & 1 x 1 & 1 x 1 & 0 x 0 \\ \hline
		B4-ReLu & 64 x 64 x 68 & - & - & - \\ \hline
		\textbf{Output : } ${H_{k}}$ & 64 x 64 x 68 & - & - & - \\ \hline
		\textbf{${L_{2}}$} \textbf{regression loss} &  &  &  &  \\ \hline
	\end{tabular}%
\end{table}

\subsection{Cascaded regression CNN}

\label{CRCNN}

The Cascaded regression CNN (CRCNN) is applied to refine the robust, but
coarse landmark estimate computed by the CHCNN subnetwork. Similar to the
CHCNN subnetwork detailed in Section \ref{Heat-map_cascaded_level}, the
CRCNN comprises of $K=4$ subunits $\left\{ \mathbf{E}_{k}\right\} _{1}^{K}$
detailed in Table \ref{table:CRCNN}. Each subunit $\mathbf{E}_{k}$ is made
of two succeeding subnetworks: the first computes a feature map of the
regression CNN using layers C1-C3 in Table \ref{table:CRCNN}, while the
second subnetwork, layers C4-C6 in Table \ref{table:CRCNN}, estimates the
residual localization error. The input $x$ to each regression subunit $%
\mathbf{E}_{k}$ is given by 
\begin{equation}
x=F_{1}\oplus F_{2}\oplus H_{k}\oplus H_{E},
\end{equation}%
that is a concatenation of both feature maps $F_{1}$ and $F_{2}$ computed by
the Base CNNs, the corresponding heatmap estimate $H_{k}$ and a baseline
heatmap estimate $H_{0}$. The output of the regression subunit $\mathbf{E}%
_{k}$ is the refinement term $\Delta \widehat{\mathbf{P}}_{k}$ as in Eq. \ref%
{equ:regression}, that is the refinement of the heatmap-based localization.
It is trained using a $L_{2}$ regression loss, and the final localization
output is given by the output of the last unit $\mathbf{E}_{K}$. 
\begin{table}[tbh]
	\caption{The landmark regression subunit. The Cascaded regression CNN
		(CRCNN) is a cascaded CNN consisting of a series of $K=4$ subunits. The
		input to each subunit is the output of the previous regression subunit, the
		corresponding heatmap unit, as well as the feature maps $F_{1}$ and $F_{2}$.}
	\label{table:CRCNN}\centering
	\par
	\begin{tabular}{|l|c|c|l|c|}
		\hline
		$\mathbf{L_{Type}}$ & \textbf{Feature Map} & $\mathbf{F_{Size}}$ & \textbf{%
			Stride} & \textbf{Pad} \\ \hline\hline
		Input : &  &  &  &  \\ 
		$F_{1}\oplus F_{2}\oplus H_{k}\oplus H_{E}$ & 64 x 64 x 332 & - & - & - \\ 
		\hline
		C1-Conv & 64 x 64 x 332 & 7 x 7 & 2 x 2 & 5 x 5 \\ \hline
		C1-Pool & 32 x 32 x 64 & 2 x 2 & 1 x 1 & 1 x 1 \\ \hline
		C2-Conv & 32 x 32 x 64 & 5 x 5 & 2 x 2 & 3 x 3 \\ \hline
		C2-Pool & 16 x 16 x 128 & 2 x 2 & 1 x 1 & 1 x 1 \\ \hline
		C3-Conv & 16 x 16 x 128 & 3 x 3 & 2 x 2 & 1 x 1 \\ \hline
		C3-Pool & 8 x 8 x 256 & 2 x 2 & 1 x 1 & 1 x 1 \\ \hline
		\textbf{\ Output :} $E_{k}$ & 8 x 8 x 256 & - & - & - \\ \hline
		\textbf{\ Input :} $E_{k}\oplus E_{k-1}$ & 8 x 8 x 512 & - & - & - \\ \hline
		C4-Conv & 8 x 8 x 512 & 3 x 3 & 2 x 2 & 1 x 1 \\ \hline
		C4-Pool & 4 x 4 x 512 & 2 x 2 & 1 x 1 & 1 x 1 \\ \hline
		C5-Conv & 4 x 4 x 512 & 3 x 3 & 2 x 2 & 1 x 1 \\ \hline
		C5-Pool & 2 x 2 x 1024 & 2 x 2 & 1 x 1 & 1 x 1 \\ \hline
		C6-Conv & 2 x 2 x 1024 & 1 x 1 & 1 x 1 & 0 x 0 \\ \hline
		\textbf{\ Output :} $\Delta \widehat{P}_{k}$ & 1 x 1 x 136 & - & - & - \\ 
		\hline
		$L_{2}$ regression loss &  &  &  &  \\ \hline
	\end{tabular}%
\end{table}

\subsection{Discussion}

The heatmap-based representation of the facial landmarks is essentially a
general-purpose metric-space representation of a set of points. The use of
smoothing filters applied to such representation relates to applying kernels
to characterize a data point based on the geometry of the points in its
vicinity \cite{COIFMAN20065,5661779}, where the use of filters of varying
support allows approximate diffusion-like analysis at different scales.
Moreover, applying multiple convolution layers and nonlinear activation
functions to the heatmaps allows to utilize convolution kernels that might
differ significantly from classical pre-designed kernels, such as Diffusion
Kernels \cite{COIFMAN20065}, as the filters in CNN-based schemes are
optimally learnt given an appropriate loss.

In the proposed scheme the heatmap is used as a state variable that is
initiated by the Base subnetwork (Section \ref{Base_Blocks}) and iteratively
refined by using two complementary losses: the heatmap-based (Section \ref%
{Heat-map_cascaded_level}) that induces the graph structure of the detected
landmarks, and the coordinates-based representation, refined by pointwise
regression (Section \ref{CRCNN}).

Such approaches might pave the way for other localization problems such as
sensor localization \cite{7055876} where the initial estimate of the heatmap
is given by a graph algorithm, rather than image domain convolutions, but
the succeeding CNN architecture would be similar to the CHCNN and CRCNN
subnetworks, and we reserve such extensions to future work.

\section{Experimental Results}

\label{results}

The proposed CCNN scheme was experimentally evaluated using multiple
contemporary image datasets used in state-of-the-art schemes, that differ
with respect to the appearance and acquisition conditions of the facial
images. We used the \textit{LFPW} \cite{Belhumeur13}, M2VTS \cite{Messer03}, 
\textit{Helen} \cite{Le2012}, AFW \cite{Ramanan12}, i-bug \cite{Sagonas16},
COFW \cite{Burgos13}, 300-W \cite{Sagonas13} and the MENPO challenge dataset 
\cite{Zafeiriou17}.

In order to adhere to the state-of-the-art 300-W competition guidelines \cite%
{Trigeorgis16,Tuzel16} $N=68$ landmarks were used in all of our experiments,
where the input RGB images were resized to $256\times 256$ dimensions, and
the pixel values were normalized to $[-0.5,0.5]$. The heatmaps were computed
at a $64\times 64$ spatial resolution, where the landmark's labeling was
applied using a symmetric Gaussian, with $\sigma =1.3$. The convolution
layers of the CCNN were implemented with a succeeding batch normalization
layer, and the training images were augmented by color changes, small angles
rotation, scaling, and translations. The learning rate was changed manually
and gradually, starting with $10^{-5}$ for the initial $30$ epochs, followed
by $5\cdot 10^{-6}$ for the next five epochs, and was then fixed at $10^{-6}$
for the remainder of the training, where the CCNN was trained for 2500
epochs.

The localization accuracy \textit{per single face image} was quantified by
the Normalized Localization Error (NLE) between the localized and
ground-truth landmarks%
\begin{equation}
NLE=\frac{1}{N\cdot d}\sum\limits_{i=1}^{N}\left\Vert \widehat{\mathbf{p}}%
_{i}-\mathbf{p}_{i}\right\Vert _{2}  \label{equ:nlm}
\end{equation}%
where $\widehat{\mathbf{p}}_{i}$ and $\mathbf{p}_{i}$ are the estimated and
ground-truth coordinates, of a particular facial landmark, respectively. The
normalization factor $d$ is either the inter-ocular distance (the distance
between the outer corners of the eyes) \cite{Ren14,Zafeiriou17,Zhu15}, or
the inter-pupil distance (the distance between the eye centers) \cite%
{Trigeorgis16}.

The localization accuracy of \textit{a set} of images was quantified by the
average localization error and the failure rate, where we consider a
normalized point-to-point localization error greater than 0.08 as a failure 
\cite{Trigeorgis16}. We also report the area under the cumulative error
distribution curve (AUC$_{\alpha }$) \cite{Trigeorgis16,Tuzel16}, that is
given by the area under the cumulative distribution summed up to a threshold 
$\alpha $. The proposed CCNN scheme was implemented in Matlab and the
MatConvNet-1.0-beta23 deep learning framework \cite{Vedaldi15} using a Titan
X (Pascal) GPU.

Where possible, we \textit{quote} the results reported in previous
contemporary works, as most of them were derived using the 300-W competition
datasets, where both the dataset and evaluation protocol are clearly
defined. In general, we prefer such a comparative approach to implementing
or training other schemes, as often, it is difficult to achieve the reported
results, even when using the same code and training set.

\subsection{300-W results}

\label{300-W-results}

We evaluated the proposed CCNN approach using the 300-W competition dataset 
\cite{Sagonas16} that is a state-of-the-art face localization dataset of $%
3,837$ near frontal face images. It comprises of images taken from the 
\textit{LFPW}, \textit{Helen}, AFW, i-bug, and \textquotedblleft 300W
private test set\textquotedblright \footnote{%
	The \textquotedblleft 300W private test set\textquotedblright\ dataset was
	originally a private and proprietary dataset used for the evaluation of the
	300W challenge submissions.} datasets. Each image in these datasets was
re-annotated in a consistent manner with $68$ landmarks and a bounding box
per image was estimated by a face detector.

The CCNN was trained using the 300-W training set and the frontal face
images of the Menpo dataset \cite{Zafeiriou17} that were annotated by 68
landmark points, same as in the 300-W dataset. The profile faces in the
Menpo dataset were annotated by 39 landmark points that do not correspond to
the 68 landmarks annotations, and thus could not be used in this work. The
overall training set consisted of $11,007$\textbf{\ }images.

The validation set was a subset of $2500$ images randomly drawn from the
training set. The face images were extracted using the bounding boxes given
in the 300-W challenge, where the shorter dimension was extended to achieve
rectangular image dimensions, and the images were resized to a dimension of $%
256\times 256$ pixels.

\subsubsection{300-W \textit{public} testset}

We compared the CCNN to contemporary state-of-the-art approaches using the 
\textit{Public} and \textit{Private} 300-W test-sets. The \textit{Public }%
test-set was split into three test datasets following the split used in
contemporary works \cite{Kowalski17,7299048}. First, the \textit{Common}
subset consisting of the test-sets of the \textit{LFPW} and \textit{Helen}
datasets (554 images overall). Second, the \textit{Challenging} subset made
of the i-bug dataset (135 images overall), and last, the 300-W public
test-set (\textit{Full Set}). The localization results of the other schemes
in Tables \ref{tab:300-W-results}-\ref{tab:Comparison_AUC} are quoted as
were reported by their respective authors.

The results are reported in Table \ref{tab:300-W-results}, and it follows
that the proposed CCNN scheme compared favorably with all other scheme,
outperforming other schemes in three out of the six test configurations. In
particular, the proposed scheme outperforms all previous approaches when
applied to the \textit{Challenging} set that is the more difficult to
localize. 
\begin{table}[tbh]
	\caption{Facial landmarks localization results of the 300-W \textit{Public}
		dataset. We report the Normalized Localization Error (NLE) as a percentage
		using the 300-W public test set and its subsets. The best results are marked
		bold.}
	\label{tab:300-W-results}\centering%
	\begin{tabular}{|c|c|c|c|}
		\hline
		\textbf{Method} & \textbf{Common set} & \textbf{Challenging set} & \textbf{%
			Full Set} \\ \hline
		\multicolumn{4}{|c|}{\textbf{\ Inter-pupil normalization}} \\ \hline
		\multicolumn{1}{|l|}{LBF\cite{Ren14}} & 4.95 & 11.98 & 6.32 \\ \hline
		\multicolumn{1}{|l|}{CFSS\cite{Zhu15}} & 4.73 & 9.98 & 5.76 \\ \hline
		\multicolumn{1}{|l|}{TCDCN\cite{Zhang16}} & 4.80 & 8.60 & 5.54 \\ \hline
		\multicolumn{1}{|l|}{RAR\cite{Xiao16}} & 4.12 & 8.35 & 4.94 \\ \hline
		\multicolumn{1}{|l|}{DRR\cite{Lai17}} & 4.07 & 8.29 & 4.90 \\ \hline
		\multicolumn{1}{|l|}{Shao et al.\cite{Shao17}} & 4.45 & 8.03 & 5.15 \\ \hline
		\multicolumn{1}{|l|}{Chen et al.\cite{Chen17}} & \textbf{3.73} & 7.12 & 
		\textbf{4.47} \\ \hline
		\multicolumn{1}{|l|}{DAN\cite{Kowalski17}} & 4.42 & 7.57 & 5.03 \\ \hline
		\multicolumn{1}{|l|}{DAN-Menpo\cite{Kowalski17}} & 4.29 & 7.05 & 4.83 \\ 
		\hline
		\multicolumn{1}{|l|}{Robust FEC-CNN\cite{He17}} & - & 6.56 & - \\ \hline
		\multicolumn{1}{|l|}{\textbf{CCNN}} & 4.55 & \textbf{5.67} & 4.85 \\ \hline
		\multicolumn{4}{|c|}{\textbf{\ Inter-ocular normalization}} \\ \hline
		\multicolumn{1}{|l|}{MDM\cite{Trigeorgis16}} & - & - & 4.05 \\ \hline
		\multicolumn{1}{|l|}{k-Convuster\cite{Kowalski16}} & 3.34 & 6.56 & 3.97 \\ 
		\hline
		\multicolumn{1}{|l|}{DAN\cite{Kowalski17}} & 3.19 & 5.24 & 3.59 \\ \hline
		\multicolumn{1}{|l|}{DAN-Menpo\cite{Kowalski17}} & \textbf{3.09} & 4.88 & 
		\textbf{3.44} \\ \hline
		\multicolumn{1}{|l|}{\textbf{CCNN}} & 3.23 & \textbf{3.99} & \textbf{3.44}
		\\ \hline
	\end{tabular}%
\end{table}

\begin{figure}[tbh]
	\centering\includegraphics[width=0.5\textwidth]{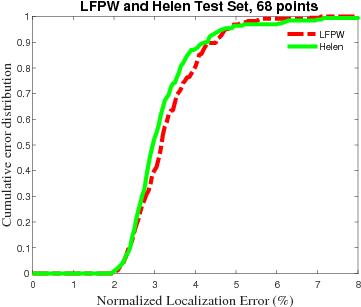}
	\caption{Facial localization results evaluated using the \textit{Helen} and 
		\textit{LFPW} testsets. We report the Cumulative Error Distribution (CED)
		vs. the normalized localization error.}
	\label{fig:Helen and LFPW}
\end{figure}

We also depict in Fig. \ref{fig:Helen and LFPW} the AUC$_{0.08}$ accuracy
measure of the CCNN when applied to the \textit{Helen} and \textit{LFPW}
testsets. 

\subsubsection{300-W \textit{private} testset}

\label{300-W private testset}

We studied the localization of the 300W \textit{Private} test set, \textit{%
	LFPW} and \textit{Helen} datasets in Table \ref{tab:Comparison_Helen_LFPW}
where the proposed scheme prevailed in four out of the six test
configurations. 
\begin{table}[tbh]
	\caption{Localization results for the LFPW, Helen and 300-W \textit{Private}
		Set. We report the Normalized Localization Error (NLE) as a percentage,
		where the best results are marked bold.}
	\label{tab:Comparison_Helen_LFPW}\centering%
	\begin{tabular}{|c|c|c|c|}
		\hline
		\textbf{Method} & \textbf{LFPW} & \textbf{Helen} & \textbf{300-W Private Set}
		\\ \hline
		\multicolumn{4}{|c|}{\textbf{\ Inter-pupil normalization}} \\ \hline
		\multicolumn{1}{|l|}{CFSS\cite{Zhu15}} & 4.87 & 4.63 & - \\ \hline
		\multicolumn{1}{|l|}{TCDCN\cite{Zhang16}} & - & 4.60 & - \\ \hline
		\multicolumn{1}{|l|}{DRR\cite{Lai17}} & \textbf{4.49} & \textbf{4.02} & - \\ 
		\hline
		\multicolumn{1}{|l|}{\textbf{CCNN}} & 4.63 & 4.51 & \textbf{4.74} \\ \hline
		\multicolumn{4}{|c|}{\textbf{\ Inter-ocular normalization}} \\ \hline
		\multicolumn{1}{|l|}{RAR\cite{Xiao16}} & 3.99 & 4.30 & - \\ \hline
		\multicolumn{1}{|l|}{MDM\cite{Trigeorgis16}} & - & - & 5.05 \\ \hline
		\multicolumn{1}{|l|}{DAN\cite{Kowalski17}} & - & - & 4.30 \\ \hline
		\multicolumn{1}{|l|}{DAN-Menpo\cite{Kowalski17}} & - & - & 3.97 \\ \hline
		\multicolumn{1}{|l|}{GAN\cite{Chen_arxiv17}} & - & - & 3.96 \\ \hline
		\multicolumn{1}{|l|}{\textbf{CCNN}} & \textbf{3.30} & \textbf{3.20} & 
		\textbf{3.33} \\ \hline
	\end{tabular}%
\end{table}

The AUC$_{0.08}$ measure and the localization failure rate are studied in
Table \ref{tab:Comparison_AUC}, where we compared against contemporary
schemes using the 300-W public and private test sets. It follows that the
proposed CCNN scheme outperforms all other schemes in all test setups. 
\begin{table}[tbh]
	\caption{AUC and failure rate of the face alignment scheme applied to the
		300-W Public and Private test sets.}
	\label{tab:Comparison_AUC}\centering%
	\begin{tabular}{|c|c|c|c|}
		\hline
		\textbf{Test Set} & \textbf{Method} & \textbf{AUC$_{0.08}$} & \textbf{%
			Failure (\%)} \\ \hline
		\multicolumn{4}{|c|}{\textbf{\ Inter-ocular normalization}} \\ \hline
		\multirow{7}{*}{\textbf{300-W Public} } & \multicolumn{1}{|l|}{ESR\cite%
			{Cao12}} & 43.12 & 10.45 \\ \cline{2-4}
		& \multicolumn{1}{|l|}{SDM\cite{Xiong13}} & 42.94 & 10.89 \\ \cline{2-4}
		& \multicolumn{1}{|l|}{CFSS\cite{Zhu15}} & 49.87 & 5.08 \\ \cline{2-4}
		& \multicolumn{1}{|l|}{MDM\cite{Trigeorgis16}} & 52.12 & 4.21 \\ \cline{2-4}
		& \multicolumn{1}{|l|}{DAN\cite{Kowalski17}} & 55.33 & 1.16 \\ \cline{2-4}
		& \multicolumn{1}{|l|}{DAN-Menpo\cite{Kowalski17}} & 57.07 & \textbf{\ 0.58}
		\\ \cline{2-4}
		& \multicolumn{1}{|l|}{\textbf{CCNN}} & \textbf{\ 57.88} & \textbf{\ 0.58 }
		\\ \hline
		\multirow{6}{*}{\textbf{300-W Private} } & \multicolumn{1}{|l|}{ESR\cite%
			{Cao12}} & 32.35 & 17.00 \\ \cline{2-4}
		& \multicolumn{1}{|l|}{CFSS\cite{Zhu15}} & 39.81 & 12.30 \\ \cline{2-4}
		& \multicolumn{1}{|l|}{MDM\cite{Trigeorgis16}} & 45.32 & 6.80 \\ \cline{2-4}
		& \multicolumn{1}{|l|}{DAN\cite{Kowalski17}} & 47.00 & 2.67 \\ \cline{2-4}
		& \multicolumn{1}{|l|}{DAN-Menpo\cite{Kowalski17}} & 50.84 & 1.83 \\ 
		\cline{2-4}
		& \multicolumn{1}{|l|}{GAN\cite{Chen_arxiv17}} & 53.64 & 2.50 \\ \cline{2-4}
		& \multicolumn{1}{|l|}{\textbf{CCNN}} & \textbf{\ 58.67} & \textbf{\ 0.83}
		\\ \hline
	\end{tabular}%
\end{table}

\begin{figure}[tbh]
	\centering
	\includegraphics[width=0.5%
	\textwidth]{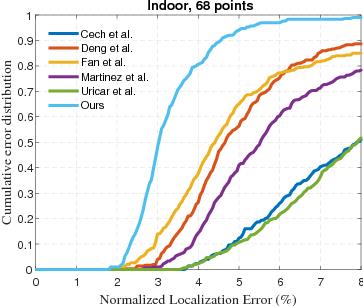}
	\caption{Facial localization results evaluated using the \textit{300-W
			Indoor }dataset. We report the Cumulative Error Distribution (CED) vs. the
		normalized localization error.}
	\label{fig:AUC_300W_Indoor}
\end{figure}

\begin{figure}[tbh]
	\centering
	\includegraphics[width=0.5%
	\textwidth]{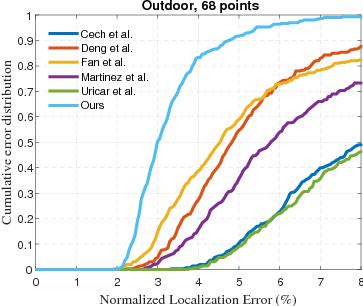}
	\caption{Facial localization results evaluated using the \textit{300-W
			Outdoor }dataset. We report the Cumulative Error Distribution (CED) vs. the
		normalized localization error.}
	\label{fig:AUC_300W_Outdoor}
\end{figure}

We also used the 300-W private testset and the corresponding split of 300 
\textit{Indoor} and 300 \textit{Outdoor} images, respectively, as these
results were reported by the prospective authors as part of the 300-W
challenge results \cite{Sagonas16}. Figures \ref{fig:AUC_300W_Indoor}-\ref%
{fig:AUC_300W_Indoor_Outdoor} depict the AUC$_{0.08}$ accuracy measure vs.
NLE, that was normalized using the inter-ocular normalization. We show the
results for the split of \textit{indoor}, \textit{outdoor} and combined (%
\textit{indoor}+\textit{outdoor}) test images in Figs. \ref%
{fig:AUC_300W_Indoor}, \ref{fig:AUC_300W_Outdoor} and \ref%
{fig:AUC_300W_Indoor_Outdoor}, respectively. The results of the schemes we
compare against are quoted from the 300-W challenge results\footnote{%
	Available at:\newline
	https://ibug.doc.ic.ac.uk/media/uploads/competitions/300w\_results.zip.}\cite%
{Sagonas16}. It follows that for all three test subsets, the proposed CCNN
scheme outperforms the contemporary schemes significantly. 
\begin{figure}[tbh]
	\centering
	\includegraphics[width=0.5%
	\textwidth]{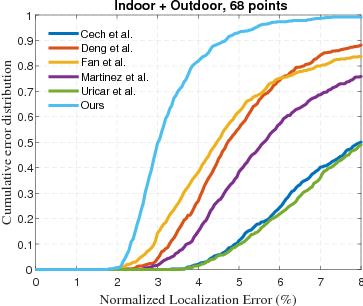}
	\caption{Facial localization results evaluated using the \textit{entire} 
		\textit{300-W }dataset. We report the Cumulative Error Distribution (CED)
		vs. the normalized localization error.}
	\label{fig:AUC_300W_Indoor_Outdoor}
\end{figure}

\begin{figure*}[tbh]
	\includegraphics[width=\textwidth]{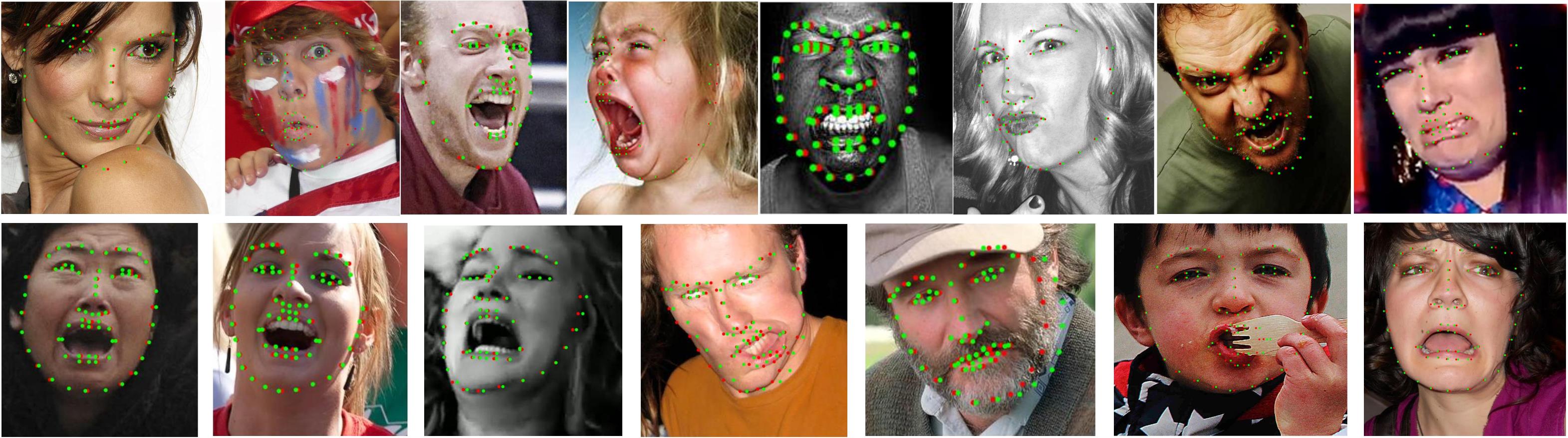}
	\caption{Facial landmarks localizations examples. The images are taken from
		the 300-W test set, where the red and green dots depict the groundtruth and
		estimated landmark points ,respectively, estimated by the proposed CCNN
		scheme.}
	\label{fig:300W_Test_Results}
\end{figure*}

Figure \ref{fig:300W_Test_Results} shows some of the estimated landmarks in
images taken from the 300-W indoor and outdoor test sets. In particular, we
show face images with significant yaw angles and facial expressions. These
images exemplify the effectiveness of the proposed CCNN framework. 

\subsection{COFW dataset results}

\label{COFW results}

\begin{figure}[tbh]
	\centering
	\includegraphics[width=0.5\textwidth]{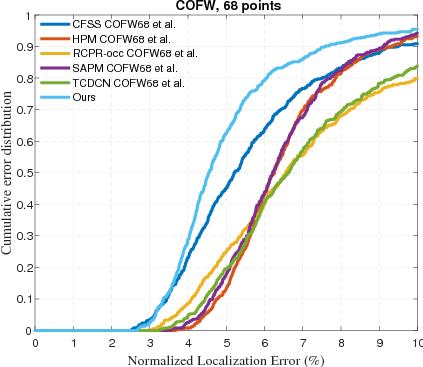}
	\caption{Facial localization results evaluated using the COFW\textit{\ }%
		dataset. We report the Cumulative Error Distribution (CED) vs. the
		normalized localization error.}
	\label{fig:AUC_COFW}
\end{figure}

The Caltech Occluded Faces in the Wild (COFW) dataset \cite{Burgos13} is a
challenging dataset consisting of $1,007$ faces depicting a wide range of
occlusion patterns, and was annotated by Ghiasi and Fowlkes \cite{Ghiasi15}
with $68$ landmark points. The common train/test split is to use 500 images
for training and the other 507 images for testing. Following previous works,
we applied the same CCNN model as in Section \ref{300-W-results} to the COFW
testset (507 images) and compared the resulting accuracy with several
state-of-the-art localization schemes. The experimental setup follows the
work of Ghiasi and Fowlkes \cite{Ghiasi15}, where the results of prior
schemes were also made public\footnote{%
	Available at: https://github.com/golnazghiasi/cofw68-benchmark}. In this
setup the CFSS \cite{Zhu15} and TCDCN \cite{Zhang16} schemes were trained
using the \textit{Helen}68, \textit{LFPW}68 and AFW68 datasets. The RCPR-occ
scheme \cite{Burgos13} was trained using the same training sets as the CCNN
model, while the HPM and SAPM schemes,\cite{Ghiasi15} were trained using 
\textit{Helen}68 and \textit{LFPW}68 datasets, respectively. The comparative
results are depicted in Fig. \ref{fig:AUC_COFW} and it follows that the CCNN
scheme outperforms the other contemporary schemes. For instance, for a
localization accuracy of 0.05 the CCNN outperforms the CFSS, coming in
second best, by close to 15\%. 

\subsection{Ablation study}

\label{Ablation results}

\begin{figure}[tbh]
	\centering\includegraphics[width=0.5\textwidth]{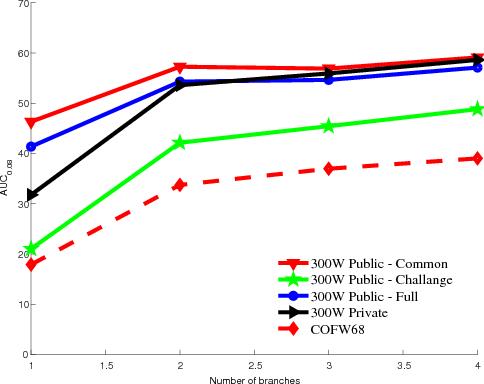}
	\caption{Ablation study results of the proposed CCNN scheme. We vary the
		number of cascades of both the CHCNN (heatmaps) and CRCNN (regression)
		subnetworks, and report the localization accuracy at the output of the CRCNN
		subnetwork.}
	\label{fig:Ablation}
\end{figure}

We studied the effectivity of the proposed CCNN cascaded architecture by
varying the number of cascades used in the CHCNN (heatmaps) and CRCNN
(regression) subnetworks. For that we trained the CCNN using $K=\left\{
1,2,3,4\right\} $ HCNN and CRCNN cascades with the same training sets and
setup as in Section \ref{300-W-results}. The resulting CNNs were applied to
the same test sets as in Sections \ref{300-W-results} and \ref{COFW results}%
. The results are depicted in Fig. \ref{fig:Ablation}, where we report the
localization accuracy at the output of the CRCNN subnetwork. It follows that
the more cascades are used the better the accuracy, and the most significant
improvement is achieved for using more than a single cascade. Moreover, it
seems that adding another cascade might improve the overall accuracy by $%
\sim 2\%$.

\section{Conclusions}

\label{conclusion}

In this work, we introduced a Deep Learning-based cascaded formulation of
the coarse-to-fine localization of facial landmarks. The proposed cascaded
CNN (CCNN) applied two dual cascaded subnetworks: the first (CHCNN)
estimates a coarse but robust heatmap corresponding to the facial landmarks,
while the second is a cascaded regression subnetwork (CRCNN) that refines
the accuracy of CHCNN landmarks localization, via regression. The two
cascaded subnetworks are aligned such that the output of each CHCNN unit is
used as an input to the corresponding CRCNN unit, allowing the iterative
refinement of the localization accuracy. The CCNN is an end-to-end solution
to the localization problem that is fully data-driven and trainable, and
extends previous results on heatmaps-based localization \cite{Belagiannis17}%
. The proposed scheme is experimentally shown to be robust to large
variations in head pose and its initialization. Moreover, it compares
favorably with contemporary face localization schemes when evaluated using
state-of-the-art face alignment datasets.

This work exemplifies the applicability of heatmaps-based landmarks
localization. In particular, the proposed CCNN scheme does not utilize any
particular appearance attribute of faces and can thus be applied, to the
localization of other classes of objects of interest. In future, we aim to
extend the proposed localization framework to the localization of sensor
networks where the image domain CNN is reformulated as graph domain CNNs.

\section{Acknowledgment}

This work has been partially supported by COST Action 1206 \textquotedblleft
De-identification for privacy protection in multimedia
content\textquotedblright\ and we gratefully acknowledge the support of
NVIDIA\textsuperscript\textregistered\ Corporation for providing the Titan X\
Pascal GPU for this research work.


\ifCLASSOPTIONcaptionsoff
\newpage
\fi

\bibliographystyle{IEEEtran}
\bibliography{IEEEabrv,mybiblio}

\end{document}